\documentclass[lettersize,journal]{IEEEtran}
\usepackage{amsmath,amsfonts,amssymb}
\usepackage{algorithmic}
\usepackage{algorithm}
\usepackage{array}
\usepackage[caption=false,font=normalsize,labelfont=sf,textfont=sf]{subfig}
\usepackage{textcomp}
\usepackage{stfloats}
\usepackage{url}
\usepackage{verbatim}
\usepackage{graphicx}
\usepackage{xcolor} 
\usepackage{cite}
\usepackage{multirow}
\usepackage{xspace}
\begin{document}

\title{Implicit Virtual Leader: Decentralized Vision-Only Relative Pose Estimation for Multi-Robot Formations}

\author{
  Shiyuan Yang$^*$, Zelin Wang$^*$, Zhijia Tao, Yilin Wang, \\
  Zhengyu Hou, Xiaosong Kong, Borong Zhang, Yip Fun Yeung, \\
  Yuankai Luo, Sharon Lee, Qingbiao Li
  \thanks{$^*$These authors contributed equally.}
}



\maketitle

\begin{abstract}
Classical leader-follower formation control suffers from single points of failure and error propagation, and relies on absolute localization sensors that are ill-suited for GPS-denied environments. We present a learned, vision-only estimator that maps each robot's monocular image, together with messages exchanged over a communication graph, directly to its 6-DoF relative pose. Its key ingredient is the implicit virtual leader (IVL): a non-physical reference frame at the team centroid, implicitly learned inside a Transformer-based graph neural network, so that estimation has no privileged node and needs no absolute localization. The estimator additionally reports well-calibrated aleatoric (heteroscedastic GNLL) uncertainty alongside epistemic (MC~Dropout) uncertainty, compared systematically across simulation and real-world test sets. Trained only in simulation, the estimator generalizes to unseen scenes, to larger unseen team sizes, and to an external real-world benchmark. It exhibits no single point of failure: removing any one robot costs at most $1.24\times$ the median removal, and removing $71\%$ of the communication links costs $1.77\times$ in position error without retraining. Trained on real-robot data from a single platform, it transfers without modification to a heterogeneous team, estimating relative pose to $0.22$\,m and $1.6^\circ$ on physical robots, where it drives closed-loop formation control.
\end{abstract}


\section{Introduction}
\IEEEPARstart{M}{ulti-robot} systems (MRS) are increasingly used to tackle complex tasks that are beyond the capabilities of a single robot, including exploration~\cite{nieto2014coordination}, search and rescue~\cite{queralta2020collaborative}, environmental monitoring~\cite{notomista2022multi}, and cargo delivery~\cite{alonso2017multi,culbertson2021decentralized,liu2023novel}. In MRS, formation control is a general capability for enabling cooperative tasks; it requires coordinating multiple robots to maintain a desired geometric configuration~\cite{abujabal2023comprehensive}. Existing strategies include leader-follower, virtual structure, and behavior-based approaches~\cite{oh2015survey}.

The leader-follower approach is popular for its simplicity, but the dependence on a designated leader introduces a single point of failure and error propagation~\cite{hu2021decentralized}. Furthermore, most formation control methods assume that each robot's absolute pose is available through GPS, LiDAR, or other dedicated sensors. These sensors can be power-hungry, bulky, and unreliable in GPS-denied or indoor environments. Monocular cameras offer a low cost alternative, but using them for pose estimation in formation control remains challenging due to sensitivity to lighting, lack of depth information, and the need for robust feature matching.

\begin{figure}[t]
\centering
\includegraphics[width=1.0\columnwidth,height=0.35\textheight,keepaspectratio]{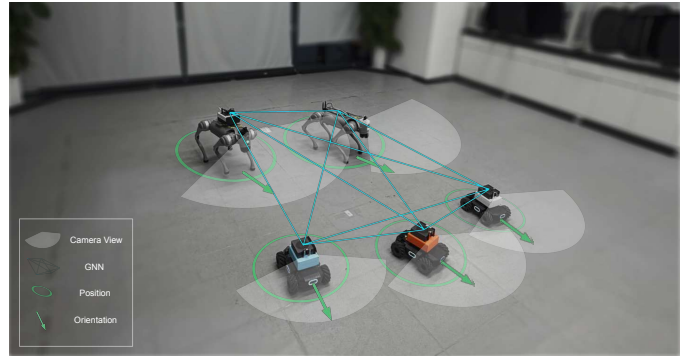}
\caption{Conceptual illustration of the proposed approach. Each robot uses only a monocular camera and inter-robot communication to estimate its relative pose with respect to an implicit virtual leader (IVL).}
\label{fig:overview}
\end{figure}

To address these challenges, we propose a decentralized relative pose estimation architecture driven solely by vision and Graph Neural Networks (GNNs). Our key idea is the implicit virtual leader (IVL): a non-physical formation reference frame that is not tied to any individual robot. Instead, it is implicitly learned within the GNN via inter-robot communication and monocular images. This eliminates both the single-point-of-failure vulnerabilities of leader-follower mechanisms and the reliance on absolute localization typical of classical virtual leader approaches. Because the estimator is learned end-to-end from images rather than derived from an analytic measurement model, it transfers without fine-tuning across team sizes, from simulation to unseen real-world scenes, and across robot platforms. We validate our method on a heterogeneous robot team comprising wheeled and quadruped platforms. The primary contributions are:

\begin{enumerate}
\item We introduce IVL, a formation reference frame implicitly learned within a GNN. Because the reference is a learned aggregate of the whole team rather than a designated robot, the estimator has no privileged node: accuracy depends only weakly on which teammate is lost without retraining.
\item We design a Transformer-based GNN that regresses each robot's pose relative to the IVL from monocular images alone, removing the reliance on absolute localization sensors or prior maps. Inference is per-node and runs onboard the robots' embedded compute in real time, at a cost that stays flat from $N=2$ to $12$.
\item We equip the IVL estimator with aleatoric uncertainty via heteroscedastic GNLL and epistemic uncertainty via MC~Dropout, and compare the two systematically. On both simulation and real-world test sets the aleatoric estimate is the better calibrated for position (ECE $0.022$ and $0.037$) and ranks errors better for both position and orientation.
\item We demonstrate sim-to-real and cross-platform transfer. The simulation trained model generalizes to unseen real-world scenes and unseen team sizes, cutting median position error by $31\%$ against the closest published baseline, at comparable orientation error. A model trained on one robot type then drives closed-loop formation control on another.
\end{enumerate}

\begin{figure*}[!t]
\centering
\subfloat[$N=2$]{\includegraphics[width=0.166\textwidth]{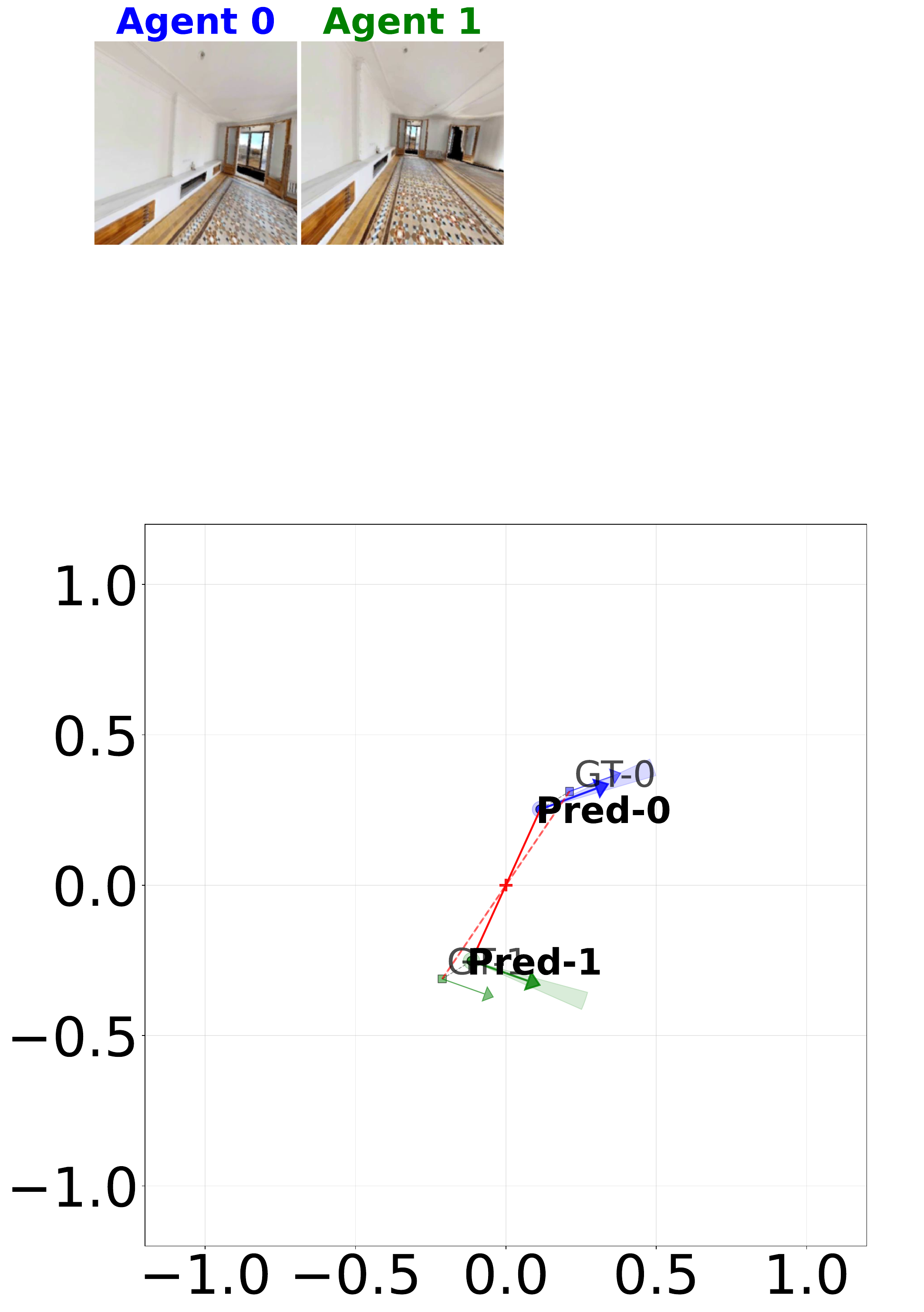}}
\hfil
\subfloat[$N=3$]{\includegraphics[width=0.166\textwidth]{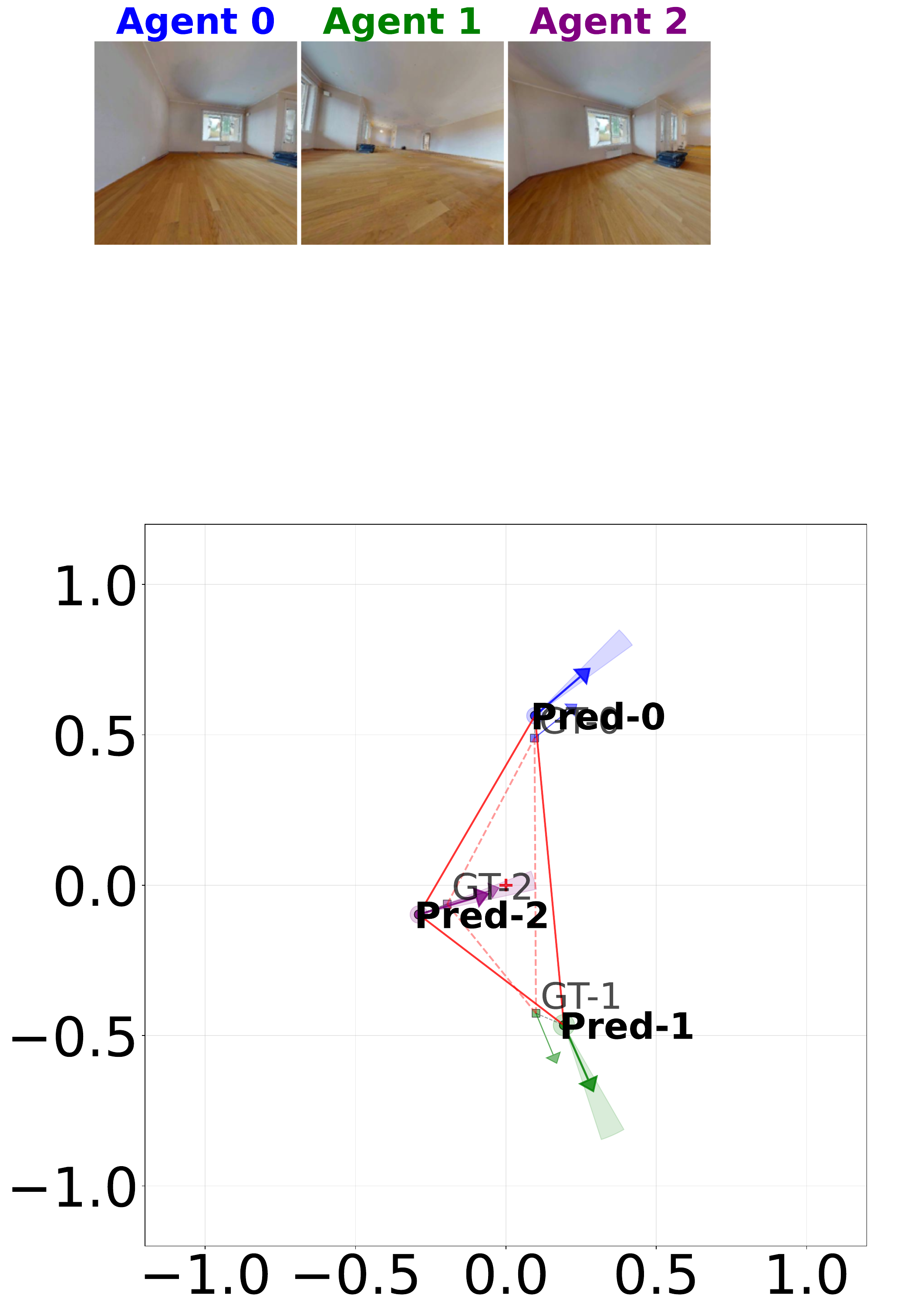}}
\hfil
\subfloat[$N=4$]{\includegraphics[width=0.166\textwidth]{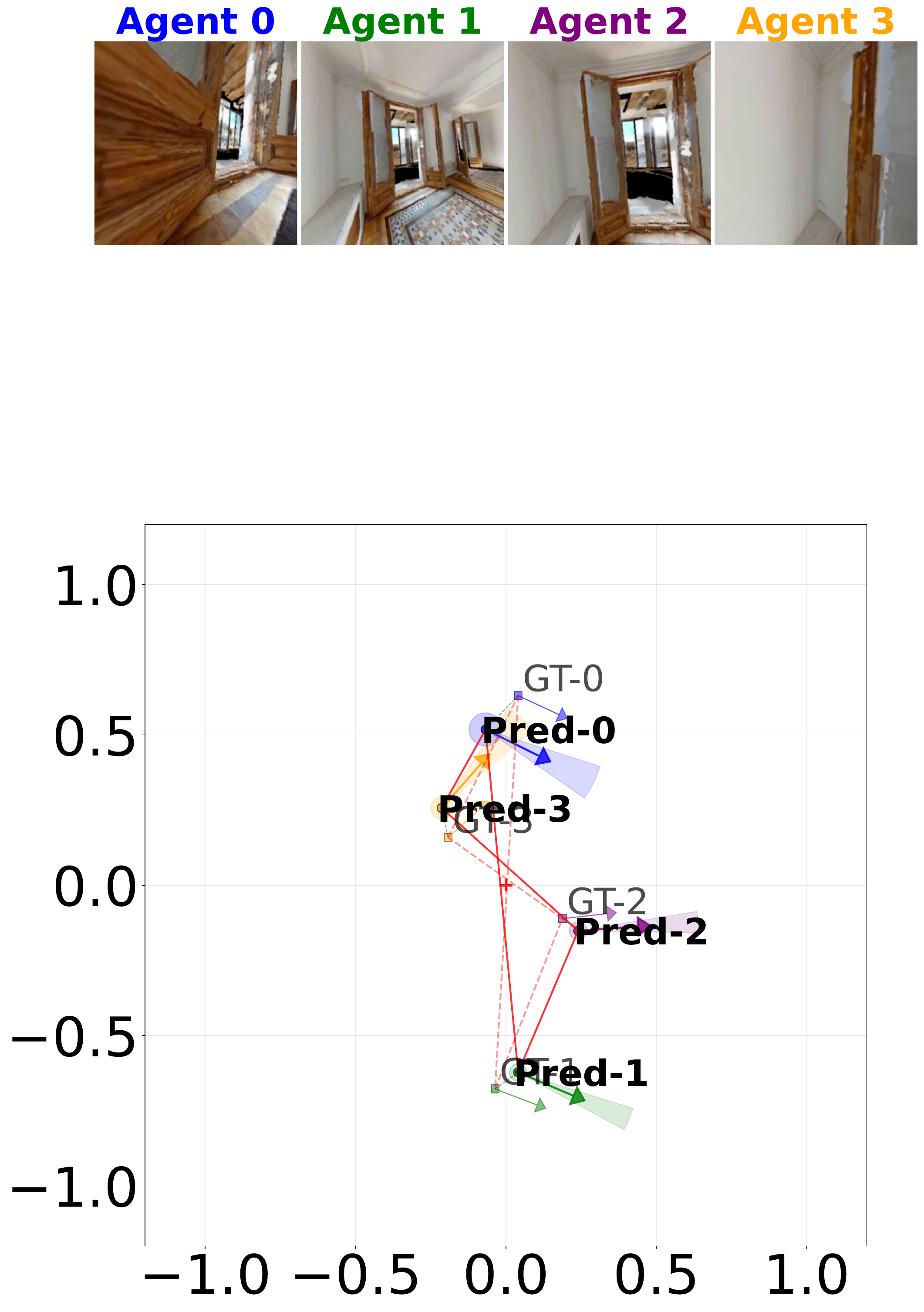}}
\hfil
\subfloat[$N=5$]{\includegraphics[width=0.166\textwidth]{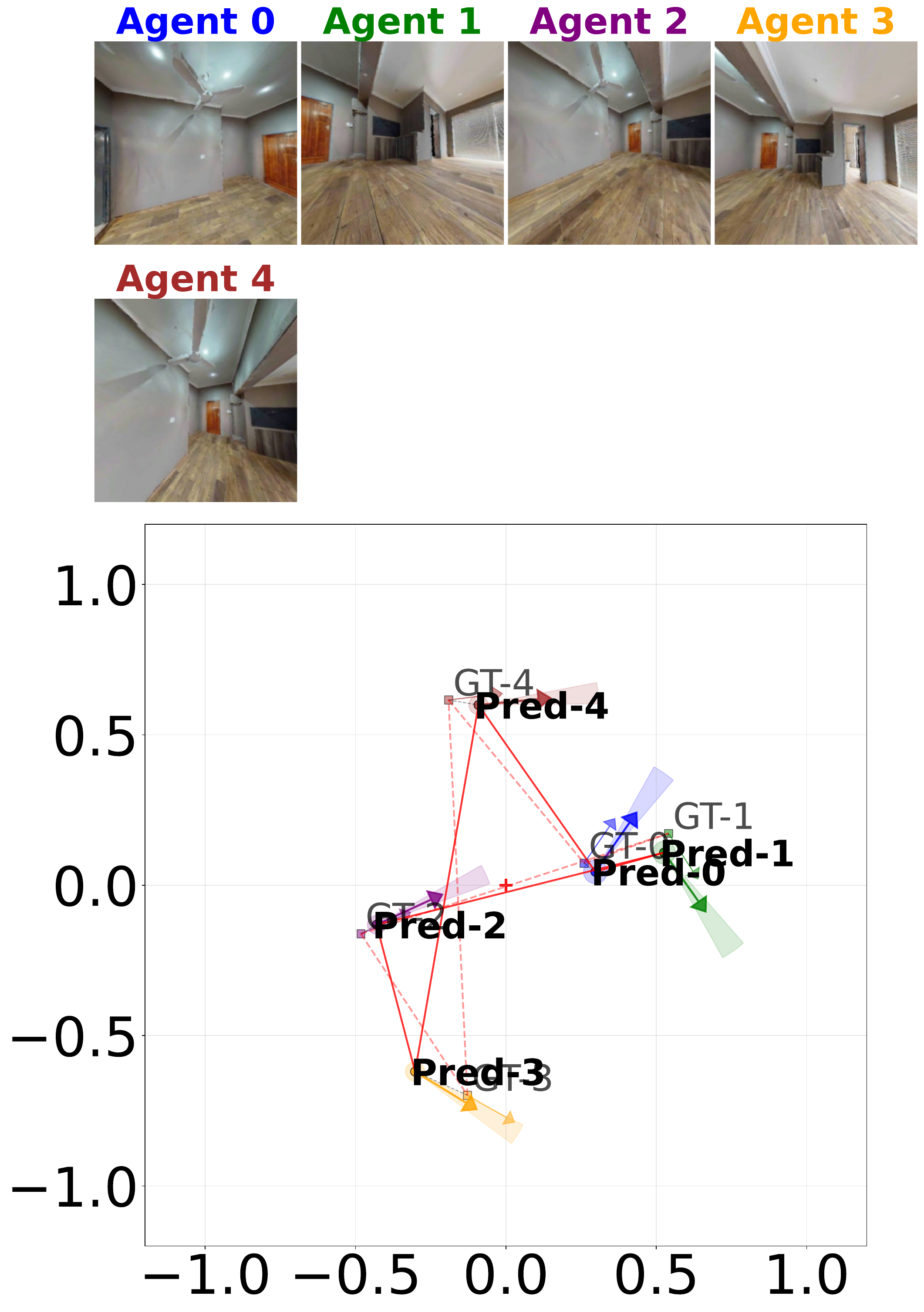}}
\hfil
\subfloat[$N=6$\textsuperscript{*}]{\includegraphics[width=0.166\textwidth]{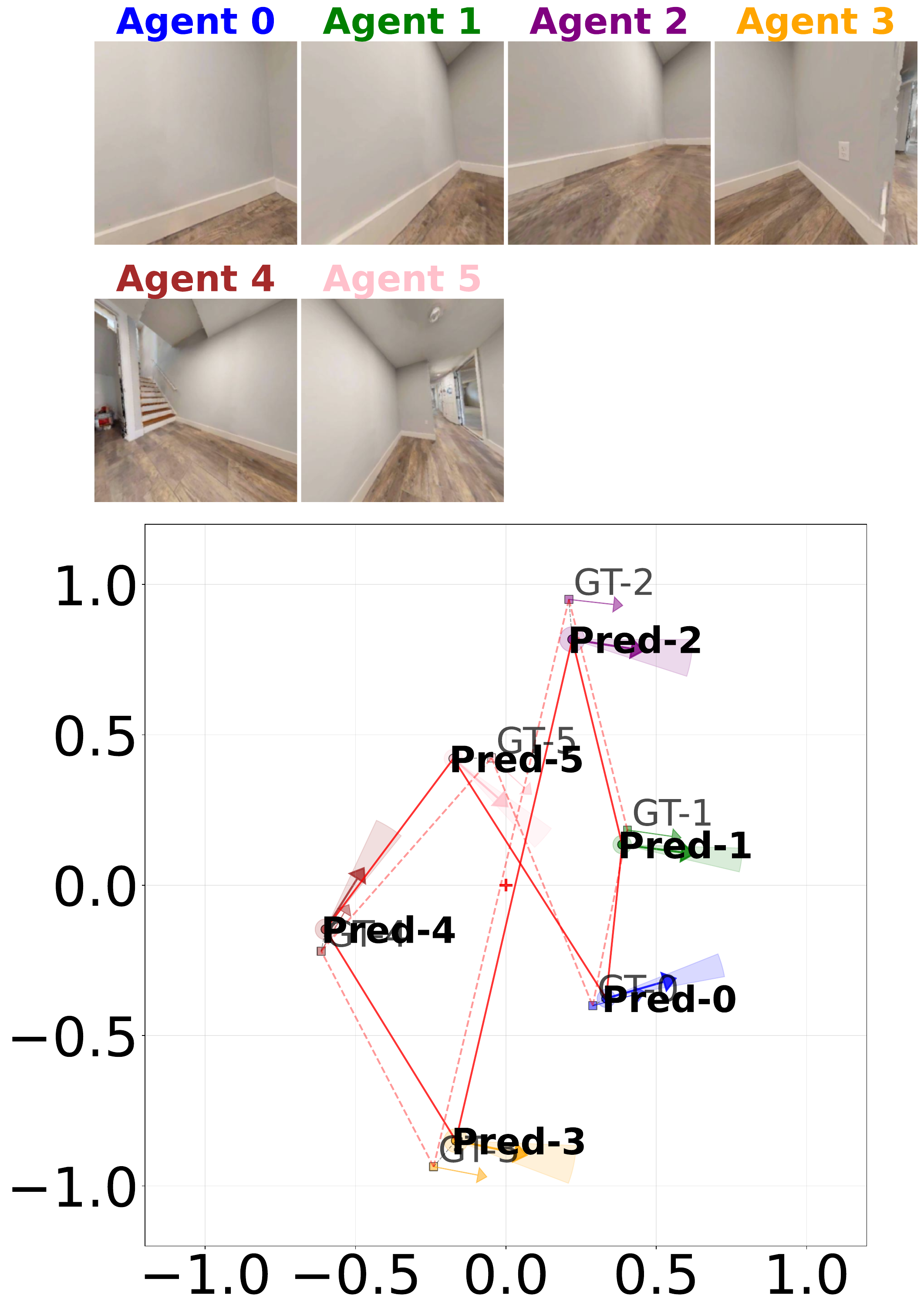}}
\hfil
\subfloat[$N=7$\textsuperscript{*}]{\includegraphics[width=0.166\textwidth]{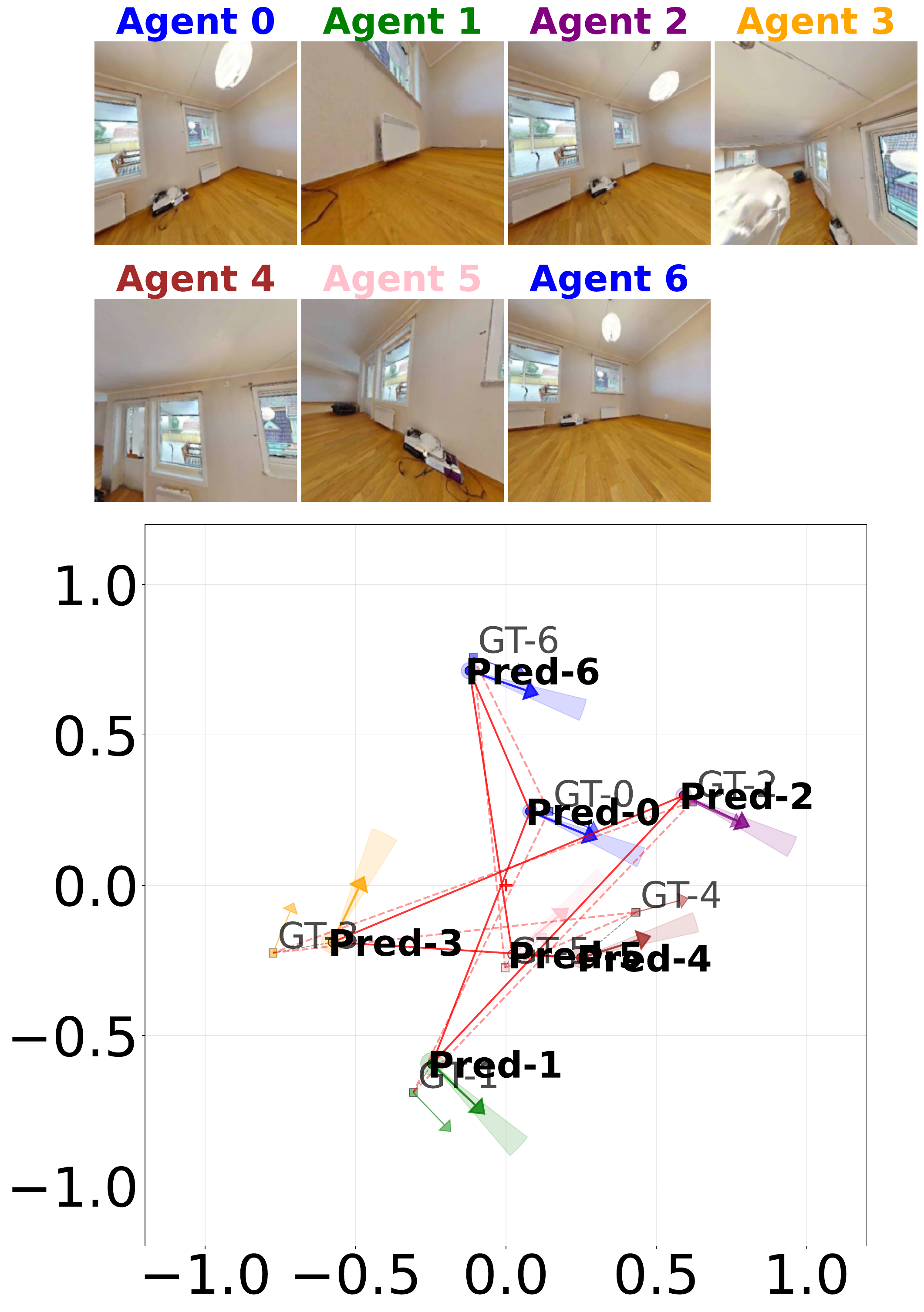}}
\caption{Qualitative pose with uncertainty visualization for team sizes $N \in \{2,3,4,5,6,7\}$. Solid markers indicate predicted poses; dashed markers indicate ground truth. Red lines connect each prediction to its ground truth position. The cross at the origin denotes the IVL. Uncertainty is visualized as position variance circles and orientation fan sectors. Asterisked sizes ($N \geq 6$) are unseen during training.}
\label{fig:qualitative}
\end{figure*}

\section{Related Work}
\noindent\textbf{\textit{Multi-robot formation control.}} Leader-follower and virtual structure paradigms have long dominated multi-robot formation control~\cite{antonelli2014decentralized,he2018leader,di2021multi,roy2018multi,tse2021relative,quan2023robust}, but both remain vulnerable in GPS-denied or cluttered environments. Leader-follower methods introduce single points of failure through leader dependence, while virtual structures often require global pose synchronization and centralized computation, limiting robustness under sensing, communication, and obstacle-induced disturbances.

\noindent\textbf{\textit{Virtual leaders.}} Researchers have proposed non-physical virtual leaders (VL) to guide robot teams~\cite{tse2021relative,leonard2001virtual,egerstedt2001formation}, mitigating the single-point-of-failure inherent to leader-follower structures. Our implicit virtual leader (IVL) extends this idea along three axes: (i) the reference frame is a statistical aggregate of the team's poses inferred from data, rather than a designer-prescribed trajectory; (ii) it is computed implicitly inside the GNN from monocular images and inter-robot embeddings, requiring no absolute pose at inference; (iii) inference is fully node-level decentralized, with no shared global state across the team.

\noindent\textbf{\textit{Decentralized cooperative estimation.}} A parallel line of work estimates robot states in a fully distributed manner by exchanging inter-robot measurements. Gaussian belief propagation (GBP) has recently emerged as a scalable peer-to-peer inference scheme: the Robot Web localizes many devices through message passing on a shared factor graph~\cite{murai2024robot}, with extensions to online SLAM with auto-calibration~\cite{murai2024distributed} and to collaborative planning~\cite{patwardhan2023distributing} that remain robust under noisy measurements and intermittent communication. Learning-based collaborative perception similarly fuses neighbor observations through graph networks for detection and tracking~\cite{zhou2022multi,song2023cooperative,lu2023robust}. These methods depend on an \emph{explicit} observation model, such as inter-robot range or bearing readings or object detections with known data association, and reason over a hand-specified factor graph. Our estimator is instead purely learned and monocular: it maps raw images and exchanged embeddings directly to metric relative pose without an analytic measurement model, and reasons over an implicit, data-driven reference rather than an externally anchored global frame.

\noindent\textbf{\textit{Vision-only swarm navigation and topology switching.}} Purely vision-based controllers can elicit collective motion in robot swarms from onboard cameras alone, without explicit communication or global positioning~\cite{mezey2025purely,bautista2024behavioral}, and formation schemes that switch topology to negotiate constrained or occluded spaces have long been studied~\cite{roy2018multi}. Such controllers typically apply hand-designed reactive rules over bearing-only or blob-level visual cues to regulate aggregate motion, rather than recovering metric relative pose. In contrast, our framework yields calibrated metric 6-DoF pose estimates precise enough to drive explicit triangle-to-column topology switching through a narrow corridor (Section~\ref{sec:tunnel}).

\noindent\textbf{\textit{Learning-based relative pose estimation with GNNs.}} GNNs have been widely adopted in multi-robot systems due to their natural ability to model communication topologies, facilitate decentralized computation, and ensure scalability~\cite{li2020graph,li2021message}, and learning-based visual relocalization has likewise matured in single-agent settings~\cite{miao2024survey}. Closest to ours, Blumenkamp et al.~\cite{blumenkamp2025covis} estimate inter-robot poses relative to a designated physical leader; this pairwise, leader-centric formulation remains susceptible to the single-point-of-failure and error-propagation issues of leader-follower schemes. Our IVL formulation instead estimates every robot's pose relative to a learned non-physical reference that is a statistical aggregate of the team, with no privileged node. Set against the paradigms above, the IVL occupies a distinct position on two axes at once. Leader-follower schemes~\cite{he2018leader}, prescribed virtual structures~\cite{leonard2001virtual} and CoViS-Net all tie their reference to something external to the estimate itself, a designated robot or a designer-specified trajectory, and so inherit an anchor. Gaussian belief propagation~\cite{murai2024robot} does dispense with the anchor, but it reasons over a hand specified factor graph fed by explicit inter-robot measurements, whereas our reference is inferred from monocular images and exchanged embeddings alone. Being simultaneously anchor free and monocular is what distinguishes the IVL, and it is what makes the robustness properties of Section~\ref{sec:robustness} available without a fallback sensor.

\section{Problem Formulation}\label{sec:problem}

We consider a team of $N$ mobile robots. Absolute poses are unavailable due to sensing or environmental constraints; each robot $i$ observes only a monocular RGB image $\mathbf{O}_i \in \mathbb{R}^{W \times H \times 3}$.
 Inter-robot communication is modeled as a directed graph $\mathcal{G}=(\mathcal{V},\mathcal{E})$, where $\mathcal{V}=\{1,\ldots,N\}$, and $\mathcal{E}$ denotes communication links. Thus, robots exchange learned feature embeddings over a communication graph within a distance threshold. We define an implicit virtual leader (IVL) $L_v=(\mathbf{p}_v,\mathbf{q}_v)$ as a non-physical reference frame associated with the formation centroid. The goal is to learn a distributed estimator $f_\theta$ that maps each robot's local observation and exchanged embeddings to its 6-DoF pose relative to the IVL, without absolute localization.

\section{Methods}

\subsection{IVL Formulation}\label{sec:ivl_formulation}
The position of the IVL is defined as the centroid of the team:
\begin{equation}
\mathbf{p}_v = \frac{1}{N}\sum_{i=1}^{N} \mathbf{p}_i.
\end{equation}
The orientation $\mathbf{q}_v$ is computed as the principal eigenvector of the quaternion correlation matrix:
\begin{equation}
\mathbf{M} = \sum_{i=1}^{N} \alpha_i\, \mathbf{q}_i \mathbf{q}_i^{\top}, \quad \mathbf{q}_v = \arg\max_{\|\mathbf{q}\|=1} \mathbf{q}^{\top} \mathbf{M} \mathbf{q},
\end{equation}
where $\alpha_1 = 1+\epsilon$ and $\alpha_{i\neq1} = 1$, with $\epsilon = 10^{-4}$. The perturbation is a numerical tie-break that resolves the degeneracy arising when several robots share the same orientation, so that the leading eigenvector is selected deterministically rather than by solver-dependent ordering. It acts only on label construction, which is performed offline from ground truth; at inference no robot is designated as a reference, which is the sense in which the estimator has no privileged node.

\begin{equation}
\mathbf{q}^{\mathrm{rel}}_i = \mathbf{q}_v^{-1} \otimes \mathbf{q}_i,
\end{equation}
\begin{equation}
\mathbf{p}^{\mathrm{rel}}_i = \mathbf{R}(\mathbf{q}_i)^{\top}(\mathbf{p}_i - \mathbf{p}_v),
\label{eq:body_frame}
\end{equation}
where $\mathbf{R}(\cdot)$ denotes the rotation matrix corresponding to a given quaternion and $\otimes$ denotes quaternion multiplication; we abbreviate $\mathbf{R}_i \triangleq \mathbf{R}(\mathbf{q}_i)$ and $\mathbf{R}_v \triangleq \mathbf{R}(\mathbf{q}_v)$ where convenient.

\begin{figure}[!t]
\centering
\includegraphics[width=\columnwidth]{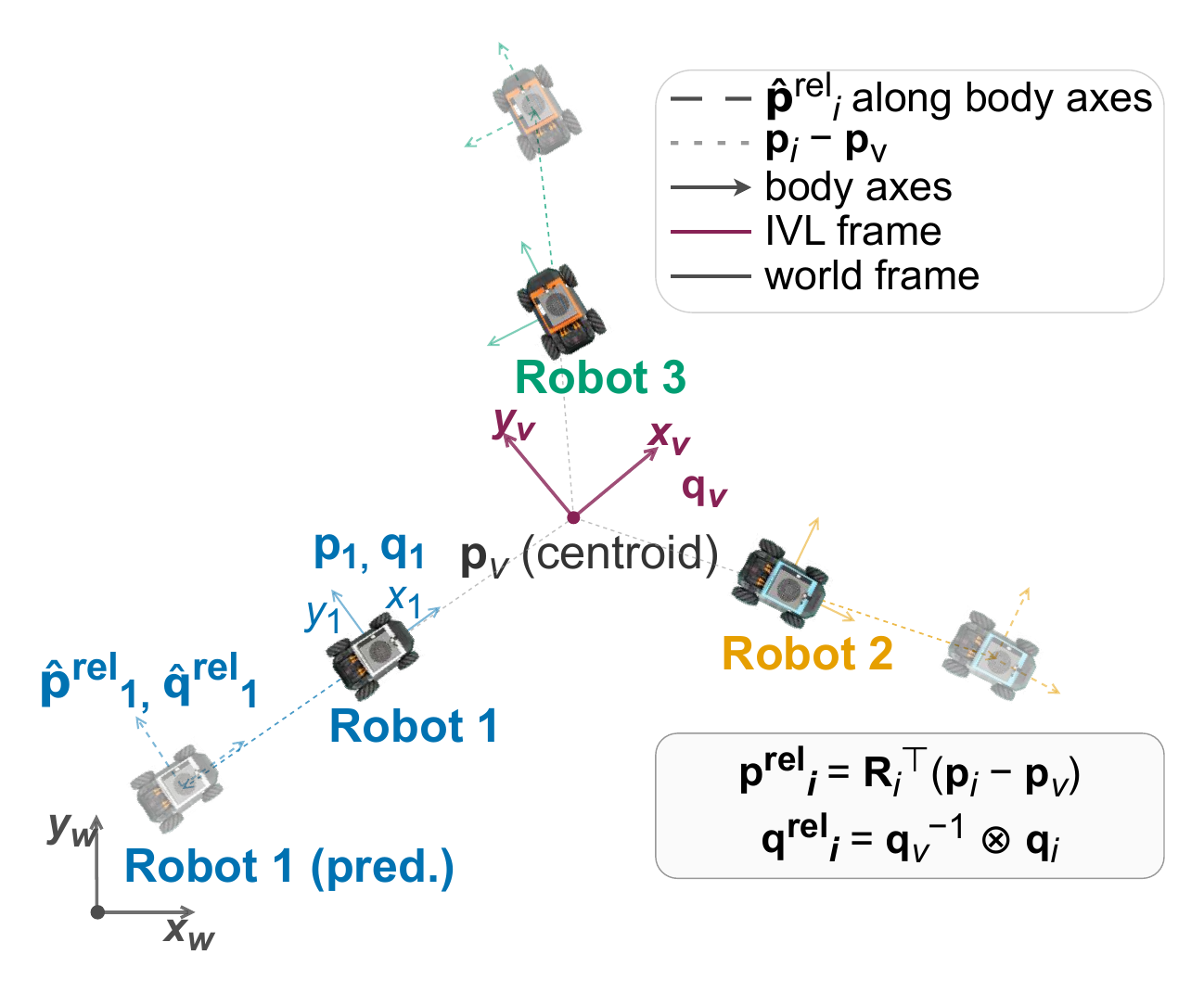}
\caption{Body frame relative position formulation. Solid robots are the true poses; faded robots are the poses predicted by the model, drawn in that robot's colour. Robot~1 is annotated in full and the same convention applies to the others. Grey dotted lines are the world-frame offsets $\mathbf{p}_i - \mathbf{p}_v$ from the centroid. The boxed equations define the ground-truth targets $\mathbf{p}^{\mathrm{rel}}_i$ and $\mathbf{q}^{\mathrm{rel}}_i$ of Eq.~\eqref{eq:body_frame}; the hatted quantities in the drawing, $\hat{\mathbf{p}}^{\mathrm{rel}}_1$ and $\hat{\mathbf{q}}^{\mathrm{rel}}_1$, are the model's estimates of them. By expressing the relative position in each robot's egocentric frame, the mapping from visual features to spatial offsets remains geometrically consistent regardless of the robot's global orientation.}
\label{fig:body_frame}
\end{figure}

The relative orientation $\mathbf{q}^{\mathrm{rel}}_i$ is expressed in the IVL frame while the relative position $\mathbf{p}^{\mathrm{rel}}_i$ is expressed in robot $i$'s body frame. This asymmetry is deliberate: orientation differences are frame invariant, so $\mathbf{q}^{\mathrm{rel}}_i$ is well defined once $\mathbf{q}_v$ is aggregated. Position offsets, in contrast, are directional vectors whose components depend on the reference frame. Expressing the position target in the IVL frame would couple the prediction frame ($\mathbf{R}_v$) with the input visual frame ($\mathbf{R}_i$) in a sample dependent way, breaking input-output equivariance and, in our experiments, causing training to stagnate. The body frame formulation in Eq.~\eqref{eq:body_frame} avoids this by mapping the centroid offset into a fixed geometry regardless of each robot's global orientation, as illustrated in Fig.~\ref{fig:body_frame}.

Algorithm~\ref{alg:body-frame} states the full construction. Beyond the definitions above it fixes two sign conventions that the equations leave open: the principal eigenvector returned by an eigensolver is determined only up to sign, so $\mathbf{q}_v$ is aligned with $\mathbf{q}_1$, and each $\mathbf{q}^{\mathrm{rel}}_i$ is mapped to the $w\geq0$ hemisphere. Without them the same formation can yield either of two opposite quaternions. The chordal loss of Eq.~\eqref{eq:chordal} is invariant to that sign, but the convention keeps the stored targets and the reported orientation metrics unambiguous.

\begin{algorithm}[t]
\caption{IVL-Referenced Relative Pose Construction}
\label{alg:body-frame}
\begin{algorithmic}[1]
\REQUIRE World poses $\{(\mathbf{p}_i, \mathbf{q}_i)\}_{i=1}^{N}$, $\mathbf{p}_i \in \mathbb{R}^3$, $\mathbf{q}_i \in \mathbb{H}$
\ENSURE Relative poses $\{(\mathbf{p}^{\mathrm{rel}}_i, \mathbf{q}^{\mathrm{rel}}_i)\}_{i=1}^{N}$
\STATE $\mathbf{p}_v \leftarrow \tfrac{1}{N}\sum_{i=1}^{N} \mathbf{p}_i$ \hfill \textit{// IVL position}
\STATE $\mathbf{M} \leftarrow \sum_{i=1}^{N} \alpha_i\, \mathbf{q}_i \mathbf{q}_i^{\top}$,\ $\alpha_1\!=\!1{+}\epsilon$,\ $\alpha_{j\neq 1}\!=\!1$ \hfill \textit{// break degeneracy}
\STATE $\mathbf{q}_v \leftarrow$ top eigenvector of $\mathbf{M}$ \hfill \textit{// IVL orientation}
\IF{$\mathbf{q}_v \cdot \mathbf{q}_1 < 0$}
    \STATE $\mathbf{q}_v \leftarrow -\mathbf{q}_v$ \hfill \textit{// sign canonicalization}
\ENDIF
\FOR{$i = 1$ \TO $N$}
    \STATE $\mathbf{q}^{\mathrm{rel}}_i \leftarrow \mathbf{q}_v^{-1} \otimes \mathbf{q}_i$
    \IF{$q^{\mathrm{rel}}_{i,w} < 0$}
        \STATE $\mathbf{q}^{\mathrm{rel}}_i \leftarrow -\mathbf{q}^{\mathrm{rel}}_i$ \hfill \textit{// $w\!\geq\!0$ hemisphere}
    \ENDIF
    \STATE $\mathbf{p}^{\mathrm{rel}}_i \leftarrow {\mathbf{R}(\mathbf{q}_i)}^{\top}(\mathbf{p}_i - \mathbf{p}_v)$
\ENDFOR
\RETURN $\{(\mathbf{p}^{\mathrm{rel}}_i, \mathbf{q}^{\mathrm{rel}}_i)\}_{i=1}^{N}$
\end{algorithmic}
\end{algorithm}

In addition to pose estimates, the model outputs associated aleatoric uncertainty: a positional variance $\hat{\boldsymbol{\sigma}}^2_{p,i} \in \mathbb{R}^3$ and an orientation variance $\hat{\sigma}^2_{q,i} \in \mathbb{R}$. The complete per robot output is an 11 dimensional vector:
\begin{equation}
\hat{\mathbf{y}}_i = [\hat{\mathbf{p}}^{\mathrm{rel}}_i, \; \hat{\boldsymbol{\sigma}}^2_{p,i}, \; \hat{\mathbf{q}}^{\mathrm{rel}}_i, \; \hat{\sigma}^2_{q,i}] \in \mathbb{R}^{11}.
\label{eq:output}
\end{equation}

\subsection{Overall Architecture}
\begin{figure*}[!t]
\centering
\includegraphics[width=\textwidth]{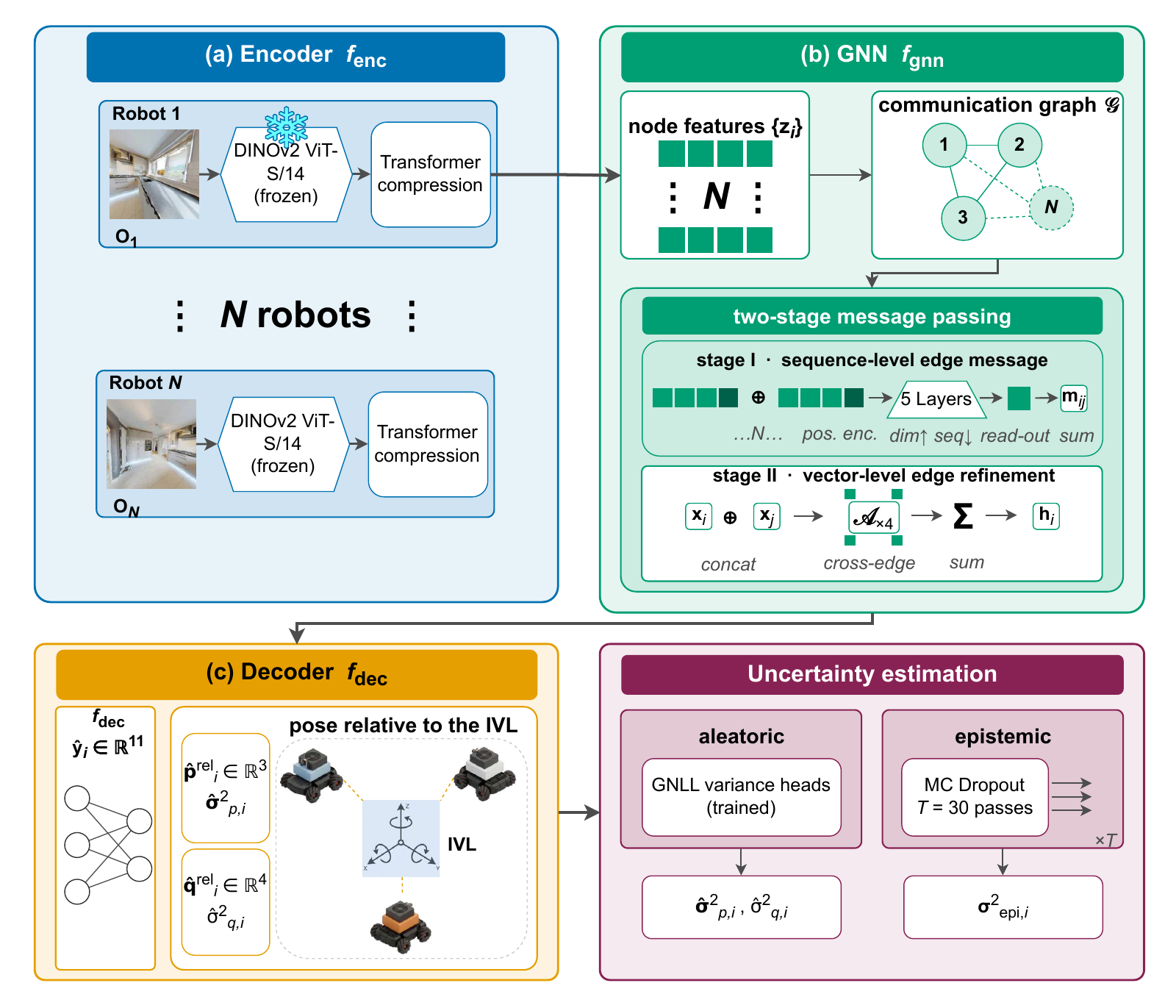}
\caption{Overview of the IVL framework. (a)~Each robot encodes its monocular observation $\mathbf{O}_i$ with a frozen DINOv2 backbone followed by a Transformer compression stage, yielding $\mathbf{z}_i \in \mathbb{R}^{L \times C}$. (b)~Two-stage Transformer message passing over the communication graph $\mathcal{G}$ forms the edge message $\mathbf{m}_{ij}$ and then the node representation $\mathbf{h}_i$. (c)~The decoder maps $\mathbf{h}_i$ to $\hat{\mathbf{y}}_i \in \mathbb{R}^{11}$, comprising the 6-DoF pose relative to the IVL and its heteroscedastic aleatoric variances. The epistemic variance $\boldsymbol{\sigma}^2_{\mathrm{epi},i}$ is obtained at inference time MC~Dropout passes.}
\label{fig:architecture}
\end{figure*}

Our method applies a GNN over the communication graph $\mathcal{G}$ defined in Section~\ref{sec:problem}. Each robot encodes its local observation, exchanges embeddings with neighbors, and decodes the aggregated node representation into its pose relative to the IVL.
Mathematically, for robot $i$, the pipeline operates as follows:
\begin{equation}
\mathbf{z}_i = f_{\mathrm{enc}}(\mathbf{O}_i),
\end{equation}
\begin{equation}
\mathbf{h}_i = f_{\mathrm{gnn}}\left(\mathbf{z}_i, \{\mathbf{z}_j \mid j \in \mathcal{N}_i\}\right),
\end{equation}
\begin{equation}
\hat{\mathbf{y}}_i = f_{\mathrm{dec}}(\mathbf{h}_i), \quad i = 1, 2, \ldots, N.
\end{equation}
Here, $f_{\mathrm{enc}}$ encodes each robot's local observation $\mathbf{O}_i$ into a feature embedding $\mathbf{z}_i \in \mathbb{R}^{L \times C}$. The GNN module $f_{\mathrm{gnn}}$ jointly processes robot $i$'s own features and those received from its neighbors $j \in \mathcal{N}_i$ via edge-level message passing, producing a compact node representation $\mathbf{h}_i$. Finally, $f_{\mathrm{dec}}$ decodes $\mathbf{h}_i$ into the estimated relative pose $\hat{\mathbf{y}}_i \in \mathbb{R}^{11}$ as defined in Eq.~\eqref{eq:output}. The proposed architecture is shown in Fig.~\ref{fig:architecture}.

\textbf{Encoder.}
The image encoder $f_{\mathrm{enc}}$ maps each robot's observation to a compact feature sequence. We use a frozen DINOv2 ViT-S/14~\cite{oquab2023dinov2} backbone followed by a multi-layer Transformer~\cite{vaswani2017attention} module that progressively reduces both the feature dimension and the sequence length. Each layer applies self-attention with residual connections, a feature-axis linear projection that progressively reduces the channel dimension, a sequence-truncation step, and a layer normalization, yielding
\begin{equation}
\mathbf{z}_i = f_{\mathrm{enc}}(\mathbf{O}_i) \in \mathbb{R}^{L \times C}, \qquad L = C = 64,
\end{equation}
where $f_{\mathrm{enc}}$ compresses the DINOv2 token tensor by an order of magnitude while preserving per-token spatial structure for downstream message passing.

\textbf{GNN.}
The GNN module $f_{\mathrm{gnn}}$ refines per-robot feature sequences $\{\mathbf{z}_i\}$ through two stage Transformer message passing on $\mathcal{G}$. First, each directed edge $(j \to i)$ processes the paired endpoint sequences to produce an edge message $\mathbf{m}_{ij} \in \mathbb{R}^{64}$, which is aggregated into an initial node embedding $\mathbf{x}_i^{(0)}$. Second, vector level edge features are jointly refined and aggregated back to each node, yielding the final representation $\mathbf{h}_i$.

\textbf{Decoder.}
The decoder regresses an 11-dimensional output for each robot:
\begin{equation}
  \hat{\mathbf{y}}_i = [\hat{\mathbf{p}}_i, \hat{\boldsymbol{\sigma}}^2_{p,i}, \hat{\mathbf{q}}_i, \hat{\sigma}^2_{q,i}],
\end{equation}
where $\hat{\mathbf{p}}_i \in \mathbb{R}^3$ and $\hat{\mathbf{q}}_i \in \mathbb{R}^4$ are the predicted relative position and L2-normalized orientation, and $\hat{\boldsymbol{\sigma}}^2_{p,i} \in \mathbb{R}^3$, $\hat{\sigma}^2_{q,i} \in \mathbb{R}$ are their associated variances.

\subsection{Uncertainty Estimation}\label{sec:uncertainty}

We model both aleatoric and epistemic uncertainty~\cite{der2009aleatory}. The aleatoric component is captured by the variance heads in Eq.~\eqref{eq:output} and trained jointly with the regression loss; the epistemic component is obtained at inference time via MC Dropout.

\subsubsection{Aleatoric Uncertainty via Heteroscedastic GNLL}

For position, we follow the heteroscedastic regression formulation of~\cite{kendall2017uncertainties} and minimize a Gaussian negative log-likelihood (GNLL) with a diagonal covariance:
\begin{equation}
\mathcal{L}_{\text{pos}} = \tfrac{1}{2}\sum_{k}\!\left(\log\hat{\sigma}_{p,i,k}^2 + \frac{(\hat{p}_{i,k} - p_{i,k}^{\text{gt}})^2}{\hat{\sigma}_{p,i,k}^2}\right),
\label{eq:lpos}
\end{equation}
where $k$ indexes the three position axes. The $\log\hat{\sigma}_{p,i,k}^2$ terms prevent the trivial solution of inflating the variances to drive the loss toward zero. For orientation, we adopt the chordal-GNLL loss~\cite{blumenkamp2025covis,peretroukhin2020smooth}, which resolves the sign ambiguity $\mathbf{q}\equiv{-}\mathbf{q}$ via:
\begin{equation}
d_{\text{quat}} = \min(\|\hat{\mathbf{q}}_i - \mathbf{q}_i^{\text{gt}}\|_2,\;\|\hat{\mathbf{q}}_i + \mathbf{q}_i^{\text{gt}}\|_2),
\label{eq:chordal}
\end{equation}
\begin{equation}
\mathcal{L}_{\text{ori}} = \tfrac{1}{2}\!\left(\log\hat{\sigma}_{q,i}^2 + \frac{2 d_{\text{quat}}^2(4 - d_{\text{quat}}^2)}{\hat{\sigma}_{q,i}^2}\right).
\label{eq:lori}
\end{equation}
At test time, the predicted variances $\hat{\boldsymbol{\sigma}}^2_{p,i}$ and $\hat{\sigma}^2_{q,i}$ are interpreted as the aleatoric uncertainty of the corresponding pose component.

\subsubsection{Epistemic Uncertainty via MC Dropout}

To capture model uncertainty, we keep dropout layers in the GNN and the pose decoder active at inference time and perform $T=30$ stochastic forward passes per input~\cite{gal2016dropout}. The DINOv2 backbone is deterministic and frozen; we cache its features once per image so each MC pass only re-evaluates the GNN and decoder, which keeps inference cost modest. The epistemic variance ($\boldsymbol{\sigma}^2_{\text{epi},i}$) is the sample variance over the $T$ predictions:
\begin{equation}
\boldsymbol{\sigma}^2_{\text{epi},i} = \frac{1}{T}\sum_{t=1}^{T}\bigl(\hat{\mathbf{y}}_i^{(t)} - \bar{\mathbf{y}}_i\bigr)^2,\quad \bar{\mathbf{y}}_i=\tfrac{1}{T}\textstyle\sum_t\hat{\mathbf{y}}_i^{(t)}.
\end{equation}
For orientation this sample variance is taken per quaternion component and then averaged, so it measures the dispersion of the predicted quaternion rather than an angular variance. It is therefore comparable to the aleatoric orientation variance in rank based terms, which is how AUSE and Spearman $\rho$ use it. We treat aleatoric and epistemic uncertainty as separate, complementary signals rather than combining them into a single scalar, as compared in Section~\ref{sec:exp_source}.

\subsection{Training Objective}

We train the network by supervised regression: at each training step, ground truth 6-DoF poses $(\mathbf{p}^{\mathrm{rel}}_i, \mathbf{q}^{\mathrm{rel}}_i)$ obtained from a simulator or mocap system serve as targets, and the network minimises the discrepancy between its predictions and those targets. The total objective combines three complementary terms:
\begin{equation}
    \mathcal{L}_{\text{total}} = \mathcal{L}_{\text{pos}} + \mathcal{L}_{\text{ori}} + \mathcal{L}_{\text{relpos}},
\end{equation}
where $\mathcal{L}_{\text{pos}}$ and $\mathcal{L}_{\text{ori}}$ are the heteroscedastic positional GNLL and chordal orientation GNLL losses defined in Section~\ref{sec:uncertainty}.

To preserve formation structure, $\mathcal{L}_{\text{relpos}}$ penalizes pairwise relative position errors in a common frame. Optimizing $\mathcal{L}_{\text{pos}}$ and $\mathcal{L}_{\text{ori}}$ independently treats each robot as a separate regression target, which can yield locally plausible poses but distort the global formation geometry. The relative position loss provides a structural constraint by matching pairwise offsets across the formation.

We lift body frame predictions to a common frame using the corresponding relative orientations:
\begin{align}
    \mathbf{w}_i^{\text{pred}} &= \mathbf{R}(\hat{\mathbf{q}}^{\mathrm{rel}}_i)\,\hat{\mathbf{p}}^{\mathrm{rel}}_i, \\
    \mathbf{w}_i^{\text{gt}}   &= \mathbf{R}(\mathbf{q}^{\mathrm{rel}}_i)\,\mathbf{p}^{\mathrm{rel}}_i,
\end{align}
and define
\begin{equation}
    \mathcal{L}_{\text{relpos}} = \frac{2}{N(N-1)}\sum_{i < j}
    \left\|
        (\mathbf{w}_i^{\text{pred}} - \mathbf{w}_j^{\text{pred}})
        -
        (\mathbf{w}_i^{\text{gt}} - \mathbf{w}_j^{\text{gt}})
    \right\|_1 .
\end{equation}
This structural term acts as a geometric regularizer, nudging the predicted formation toward the correct pairwise geometry rather than fitting each robot in isolation; Section~\ref{sec:pose_estimation_accuracy} quantifies its effect via an ablation.

\subsection{Training Details}\label{sec:training}
Excluding the frozen DINOv2 backbone, our model contains approximately 8.4\,M trainable parameters. We use the HM3D dataset, partitioned into 638 training, 152 validation, and 8 test scenes, yielding a total of 800\,k formation samples. Each sample contains a varying number of robots, with the team size $N$ randomly drawn from $\{2, \dots, 5\}$. To assemble a formation, rendered viewpoints within a scene are grouped into navigable-point clusters; each sample takes $N$ distinct clusters on the same floor with one image per cluster, and all pairwise distances stay within the $2.0$\,m formation diameter. Input images are resized to $224\times224$ pixels. Training runs for 15 epochs with a batch size of 64 on a single NVIDIA RTX 5090 GPU using bf16 mixed precision, requiring approximately 9 hours to converge. 

\section{Experimental Results}

We evaluate the proposed method to answer the following key questions:
\begin{itemize}
\item \textbf{Q1:} Does the IVL-based formulation accurately estimate relative poses across varying formation sizes?
\item \textbf{Q2:} Is the estimator robust to robot failure (node dropout) and communication loss (edge sparsification)?
\item \textbf{Q3:} Do the predicted uncertainty estimates faithfully reflect actual errors and remain reliable under sim-to-real domain shift?
\item \textbf{Q4:} Does the model generalize to entirely unseen environments without fine-tuning?
\item \textbf{Q5:} Can the system be deployed on heterogeneous robot platforms and operate in real-world formation control?
\end{itemize}

\subsection{Experimental Setup}

\textbf{1) HM3D (Habitat Matterport3D) Dataset:}
We evaluate on the 8 HM3D~\cite{ramakrishnan2021hm3d} test scenes described in Section~\ref{sec:training}, which are entirely unseen during training. Each scene provides photorealistic multi-room indoor renderings with ground truth 6-DoF camera poses used as the evaluation reference.

\textbf{2) CoViS-Net Test Set:}
We use the CoViS-Net real-world benchmark~\cite{blumenkamp2025covis}, which spans diverse indoor environments and one outdoor scene with LiDAR-based ground truth, as an out-of-distribution (OOD) test set. To assess sim-to-real generalization, the model trained in simulation is directly evaluated on the real-world dataset without any fine-tuning.

\textbf{3) Real-world Platform:}
Our real-world experimental platform comprises three modified DJI RoboMaster (RM) wheeled robots and one Unitree Go2 quadruped. Each RM is equipped with a monocular RGB camera ($148^\circ$ field of view) and an NVIDIA Jetson AGX Orin compute module. Similarly, the quadruped is retrofitted with the identical camera model, alongside a Jetson Orin Nano. Experiments are conducted in a $6.3 \times 5.1$\,m indoor laboratory arena, where ground truth 6-DoF poses are recorded using a mocap system.

\begin{table}[!t]
\caption{Pose estimation accuracy on the HM3D test split (8 scenes, $N\in\{2,3,4,5\}$, 989{,}962 samples, pooled median); samples are obtained by enumerating all $N$-subsets at each test timestamp. Position in meters, orientation in degrees. Visibility: $|\Delta\psi|<120^\circ$; filtering uses Youden's index on predicted variance $\hat{\sigma}_p$. Best per column in \textbf{bold}.}
\label{tbl:hm3d_comparison}
\centering
\setlength{\tabcolsep}{2pt}
\begin{tabular}{l|cc|cc|cc|cc}
\hline
 & \multicolumn{2}{c|}{All} & \multicolumn{2}{c|}{Visible} & \multicolumn{2}{c|}{Invisible} & \multicolumn{2}{c}{Invis.~Filt.} \\
Method & pos & ori & pos & ori & pos & ori & pos & ori \\
\hline
CoViS-Net~\cite{blumenkamp2025covis} & 0.403 & 8.3 & 0.279 & 7.5 & 0.442 & 8.5 & \textbf{0.222} & 6.8 \\
Ours                              & \textbf{0.277} & \textbf{6.7} & \textbf{0.199} & \textbf{4.6} & \textbf{0.305} & \textbf{8.4} & 0.230 & \textbf{5.8} \\
\hline
\end{tabular}
\end{table}

\begin{figure}[!t]
\centering
\includegraphics[width=\columnwidth]{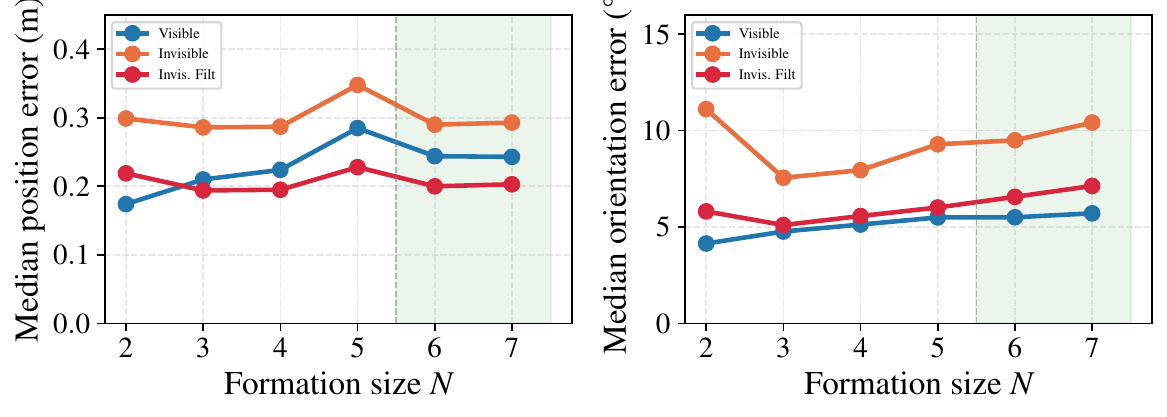}
\caption{Median pose errors across formation sizes $N \in \{2,\dots,7\}$ for four conditions: Visible, Visible+Filtered (Vis.\ Filt), Invisible, and Invisible+Filtered (Invis.\ Filt). The green shaded region marks formation sizes unseen during training ($N > 5$).}
\label{fig:realworld_conditions}
\end{figure}

\subsection{Pose Estimation Accuracy}\label{sec:pose_estimation_accuracy}
To answer Q1, we evaluate position and orientation accuracy across varying formation sizes. Since both predictions and ground truth labels are expressed as relative poses (Section~\ref{sec:ivl_formulation}), errors are computed in this relative coordinate system. The position error for robot $i$ is the Euclidean distance between the predicted and ground truth body frame offsets:
\begin{equation}
e_p^i = \|\hat{\mathbf{p}}_i^{\mathrm{rel}} - \mathbf{p}_i^{\mathrm{rel,gt}}\|_2,
\end{equation}
where $\mathbf{p}_i^{\mathrm{rel}} = \mathbf{R}_i^\top (\mathbf{p}_i - \mathbf{p}_v)$ denotes the centroid offset rotated into robot $i$'s body frame. The orientation error is the geodesic distance between the predicted and ground truth relative quaternions:
\begin{equation}
e_q^i = 2\arccos(|\langle \hat{\mathbf{q}}_i^{\mathrm{rel}}, \mathbf{q}_i^{\mathrm{rel,gt}} \rangle|),
\end{equation}
where $\mathbf{q}_i^{\mathrm{rel}} = \mathbf{q}_v^{-1} \otimes \mathbf{q}_i$, expressed in degrees. We report median errors to reduce sensitivity to outliers.

We benchmark our model against CoViS-Net~\cite{blumenkamp2025covis} on 8 indoor test scenes split from the HM3D dataset in Table~\ref{tbl:hm3d_comparison}. Both methods are trained on HM3D and evaluated on the same eight unseen scenes. We report the pooled median error over all formation sizes $N\in\{2,3,4,5\}$ across 989,962 samples. A pair of nodes is classified as visible if the two cameras share FOV overlap ($|\Delta\psi|<120^\circ$) and invisible otherwise. High uncertainty predictions are further rejected by Youden's index on the predicted variance $\hat{\sigma}_p$. As shown in Table~\ref{tbl:hm3d_comparison}, our IVL-based formulation achieves accuracy highly competitive with the CoViS-Net baseline across both visible and invisible conditions. Pooling hides the spread between scenes, so we also report it: the per-scene median position error ranges from $0.198$\,m to $0.333$\,m, and a scene-level bootstrap over the eight scenes gives a $95\%$ confidence interval of $[0.198,0.245]$\,m for the median scene. The pooled figure sits above most scenes because the hardest scene also supplies $52\%$ of the samples, so it is a conservative rather than a favourable summary. 

We further evaluate performance per formation size on HM3D across $N \in \{2, \dots, 7\}$; the model is trained on $N \in \{2, \dots, 5\}$, so $N \in \{6, 7\}$ are entirely unseen. Fig.~\ref{fig:realworld_conditions} breaks down errors by visibility and uncertainty filtering; the trends mirror Table~\ref{tbl:hm3d_comparison}. Performance remains stable on the unseen sizes: position error is 0.29~m for both $N=6$ and $N=7$, with orientation errors of 9.3$^\circ$ and 10.3$^\circ$ respectively. This demonstrates that the GNN based architecture naturally generalizes to larger teams, as each node's prediction depends only on aggregated neighbor features rather than a fixed team size.

\begin{table}[t]
\centering
\caption{Ablation of the structural relative-position loss $\mathcal{L}_{\text{relpos}}$ on the HM3D test scenes (median over 3{,}000 formations). Per-node position is the body-frame offset error; pairwise shape is the mean discrepancy of pairwise relative offsets lifted to a common frame, i.e.\ the quantity $\mathcal{L}_{\text{relpos}}$ regularizes. Lower is better.\label{tbl:relpos_ablation}}
\setlength{\tabcolsep}{4pt}
\begin{tabular}{l|cc|cc}
\hline
 & \multicolumn{2}{c|}{$N=3$} & \multicolumn{2}{c}{$N=5$} \\
Metric (m) & w/ & w/o & w/ & w/o \\
\hline
Per-node position & \textbf{0.253} & 0.258 & \textbf{0.291} & 0.298 \\
Pairwise shape    & \textbf{0.437} & 0.448 & \textbf{0.490} & 0.504 \\
\hline
\end{tabular}
\end{table}

To isolate its contribution, we train an otherwise identical variant without $\mathcal{L}_{\text{relpos}}$ and evaluate both models on the same test scenes (Table~\ref{tbl:relpos_ablation}). Removing the structural term consistently raises both the per-node position error and the pairwise formation-shape error by roughly 2--3\% across team sizes, with the shape metric, which measures formation geometry directly, affected slightly more than per-node accuracy. The structural loss thus provides a small but consistent geometric regularization; it improves formation-geometric consistency without being critical to convergence.

\subsection{Robustness to Robot Failure and Communication Loss}\label{sec:robustness}

\begin{figure*}[!t]
\centering
\includegraphics[width=\textwidth]{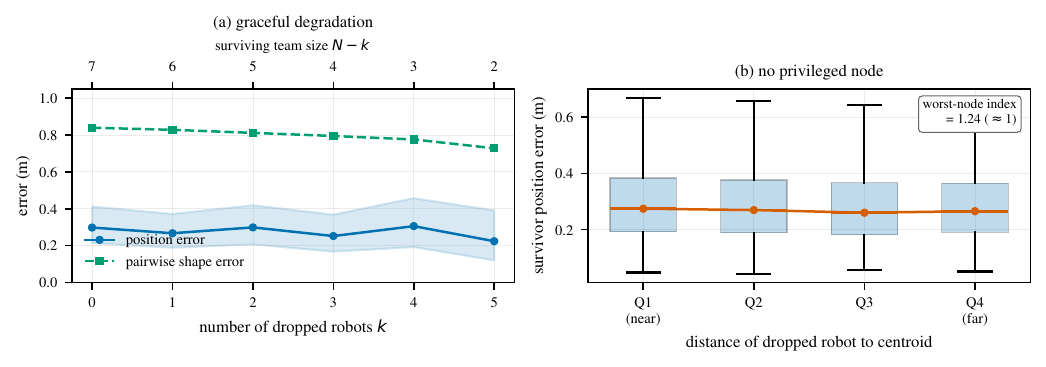}
\caption{\textbf{Node-dropout robustness on HM3D} (base $N=7$, all $\binom{7}{k}$ removals). (a) Surviving robots' error stays flat as the team shrinks: the model re-anchors to the smaller team's centroid. (b) Grouping the survivor error by how central the removed robot is (its distance to the centroid), the error depends only weakly on \emph{which} robot is lost: the quartile medians span $0.260$--$0.274$\,m and the per-formation worst-node index is $1.24$, so no single robot is critical.}
\label{fig:node_dropout}
\end{figure*}

\begin{figure}[!t]
\centering
\includegraphics[width=\columnwidth]{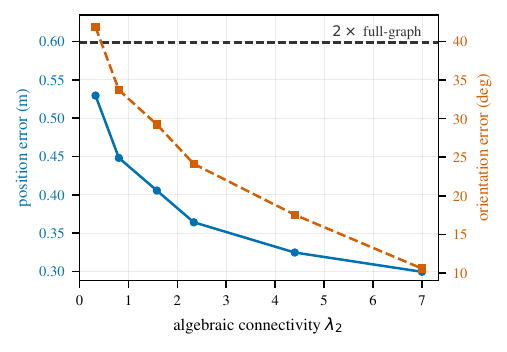}
\caption{\textbf{Edge-dropout robustness on HM3D} ($N=7$, connected graphs only). Position and orientation error degrade gracefully and monotonically with the algebraic connectivity $\lambda_2$; even the sparsest connected graph stays below $2\times$ the fully-connected position error (dashed line). Orientation degrades considerably more than position.}
\label{fig:edge_dropout}
\end{figure}

To answer Q2, we apply two structural perturbations of the running team at inference, on the HM3D test scenes, using the same model as above. Both are estimation-level stress tests; we report pooled median errors over the surviving robots.

\textbf{Node dropout.} Starting from a base formation of $N=7$ (two sizes beyond the $N\!\le\!5$ used in training, so the sweep doubles as an unseen cardinality stress test), we remove $k$ robots (enumerating all $\binom{7}{k}$ subsets, $k=1,\dots,5$, keeping a surviving team of at least two), recompute the IVL over the survivors, and measure their relative-pose error. Unlike the cardinality sweep of Fig.~\ref{fig:realworld_conditions}, where each team size is an independently sampled formation, this is a \emph{within-team} ablation that isolates the effect of losing a specific teammate. As shown in Fig.~\ref{fig:node_dropout}(a), the survivors' position error stays essentially flat as the team shrinks ($0.22$--$0.31$\,m across all $k$): the model correctly re-anchors to the smaller team's centroid, and the pairwise formation shape is preserved. More importantly, Fig.~\ref{fig:node_dropout}(b) groups the survivor error by how central the removed robot was, and finds it invariant to \emph{which} robot is lost (medians $0.274/0.270/0.260/0.266$\,m from the most central to the most peripheral quartile); the per-formation worst-node index $\max_j E(\text{drop }j)/\mathrm{median}_j E(\text{drop }j)$ is only $1.24$ ($1.20$ for the reference-invariant shape error). A leader-centric estimator has no such property, since losing the anchor forfeits the shared reference frame. Geometrically, removing one robot shifts the IVL by only $\lVert\mathbf{c}-\mathbf{p}_j\rVert/(N-1)$ (median $0.13$\,m at $k{=}1$, up to $0.48$\,m at $k{=}5$), an $O(\text{radius}/N)$ perturbation, whereas a designated leader's failure induces an $O(\text{diameter})$ re-anchoring jump: over the same $3000$ formations, re-electing the nearest surviving robot displaces the frame by a median $0.57$\,m, and by $1.42$\,m for the worst re-election, against a formation diameter of $1.92$\,m. Together these establish that the IVL has no single point of failure.

\textbf{Edge dropout.} Keeping all $N=7$ robots present, we instead sparsify the communication graph, drawing a random spanning tree and adding Bernoulli-sampled extra edges so that it stays connected, since a robot with no communication cannot participate in a formation. Although the network was trained only on fully-connected graphs, position accuracy degrades gracefully and monotonically with the graph's algebraic connectivity $\lambda_2$ (Fig.~\ref{fig:edge_dropout}): even the sparsest connected topology, a spanning tree ($\lambda_2\approx0.33$, $71\%$ of edges removed), stays at $0.53$\,m, only $1.77\times$ the fully-connected $0.30$\,m, so no retraining is required to tolerate substantial link loss. Across structured topologies the error tracks connectivity (full $<$ ring $<$ star $\approx$ chain), with one telling exception: the star ($\lambda_2{=}1.0$) is no better than the ring ($\lambda_2{=}0.75$) because each of its leaves retains only a single neighbor, indicating that the \emph{minimum} node degree, not $\lambda_2$ alone, governs per-node accuracy. Orientation degrades far more than position under sparsification (median rising from $10.6^\circ$ to $42^\circ$ at the sparsest connected graph); since inter robot clearance is set by position, position accuracy is the quantity that matters most for formation keeping. We note that this sweep is an offline, estimation level stress test over connected graphs.
\subsection{Uncertainty Estimation Evaluations}\label{sec:exp_source}

To answer Q3, we use three complementary metrics to evaluate uncertainty quality~\cite{guo2017calibration,ilg2018uncertainty}:

1)~\textbf{ECE}~\cite{guo2017calibration} (Expected Calibration Error) measures calibration fidelity by binning predictions and comparing mean variance estimate $\bar{\sigma}_b$ to mean absolute error $\bar{e}_b$ in each bin $b$:
\begin{equation}
\mathrm{ECE} = \sum_{b} \frac{|B_b|}{n} \bigl|\bar{\sigma}_b - \bar{e}_b\bigr|,
\label{eq:ece}
\end{equation}
where $|B_b|$ is the bin count and $n$ is the total number of samples. A lower ECE indicates a smaller gap between predicted confidence and actual error magnitudes, and therefore greater reliability.

2)~\textbf{AUSE}~\cite{ilg2018uncertainty} (Area Under the Sparsification Error curve) measures filtering utility: samples are progressively removed in order of decreasing uncertainty and prediction error is recomputed at each removal fraction $\theta$:
\begin{equation}
\mathrm{AUSE} = \int_0^1 \bigl[E(\theta) - E_{\mathrm{oracle}}(\theta)\bigr]\,d\theta,
\label{eq:ause}
\end{equation}
where $E(\theta)$ is the error after removing fraction $\theta$ by uncertainty and $E_{\mathrm{oracle}}(\theta)$ removes by true error. The lower the AUSE, the better uncertainty identifies unreliable predictions.

3)~\textbf{Spearman~$\rho$} measures the monotonic rank correlation between predicted uncertainty and actual error:
\begin{equation}
\rho = 1 - \frac{6 \sum_{i=1}^{n} d_i^2}{n(n^2 - 1)},
\label{eq:spearman}
\end{equation}
where $d_i$ is the difference between the rank of the predicted uncertainty $\hat{\sigma}_i$ and the rank of the actual error $e_i$, and $n$ is the total number of samples. The higher the $\rho$, the more reliable the uncertainty signal is as a ranking proxy, even when absolute magnitudes are miscalibrated.

\begin{table}[!t]
\caption{Uncertainty source comparison on the simulation ($D_{\text{Test}}^{\text{Sim}}$) and real-world ($D_{\text{Test}}^{\text{Real}}$) test sets. Best per column in \textbf{bold}. ECE: Expected Calibration Error; AUSE: Area Under Sparsification Error curve; $\rho$: Spearman rank correlation between uncertainty and error.}
\label{tbl:uncertainty_source}
\centering
\setlength{\tabcolsep}{3pt}
\begin{tabular}{l|l|c c c|c c c}
\hline
 & & \multicolumn{3}{c|}{Position} & \multicolumn{3}{c}{Orientation} \\
Setting & Source & ECE$\downarrow$ & AUSE$\downarrow$ & $\rho$$\uparrow$ & ECE$\downarrow$ & AUSE$\downarrow$ & $\rho$$\uparrow$ \\
\hline
\multirow{2}{*}{Simulation}
 & Aleatoric & \textbf{0.022} & \textbf{0.169} & \textbf{0.676} & \textbf{0.113} & \textbf{0.174} & \textbf{0.663} \\
 & Epistemic & 0.241 & 0.419 & 0.223 & 0.425 & 0.265 & 0.600 \\
\hline
\multirow{2}{*}{Real-world}
 & Aleatoric & \textbf{0.037} & \textbf{0.235} & \textbf{0.473} & \textbf{0.082} & \textbf{0.176} & \textbf{0.386} \\
 & Epistemic & 0.173 & 0.302 & 0.360 & 0.339 & 0.244 & 0.374 \\
\hline
\end{tabular}
\end{table}

\begin{figure}[!ht]
\centering
\captionsetup[subfloat]{justification=centering}
\subfloat[Position (Simulation)]{\includegraphics[width=0.48\columnwidth]{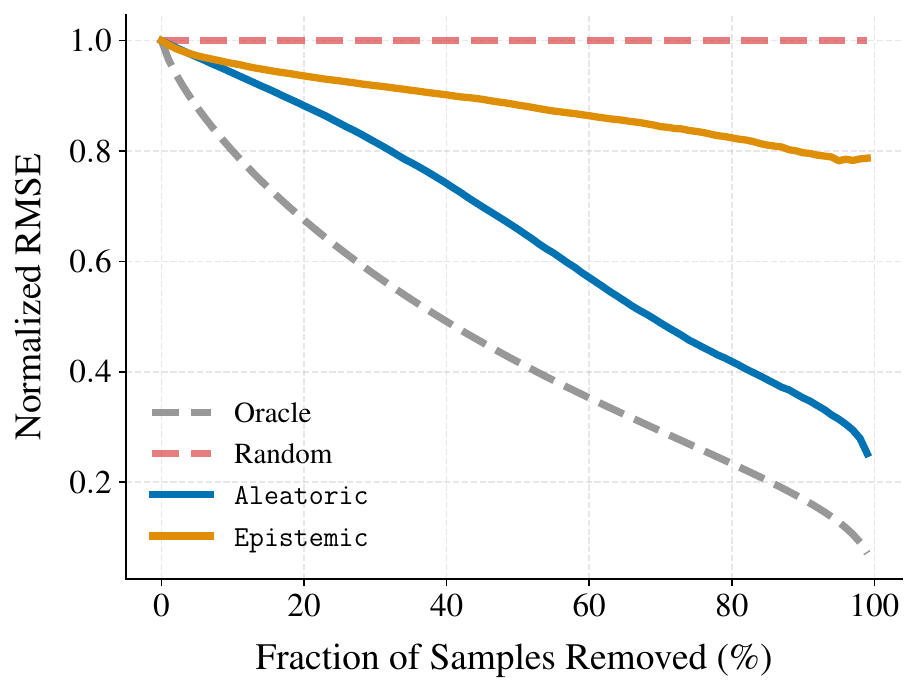}}
\hfil
\subfloat[Position (Real-world)]{\includegraphics[width=0.48\columnwidth]{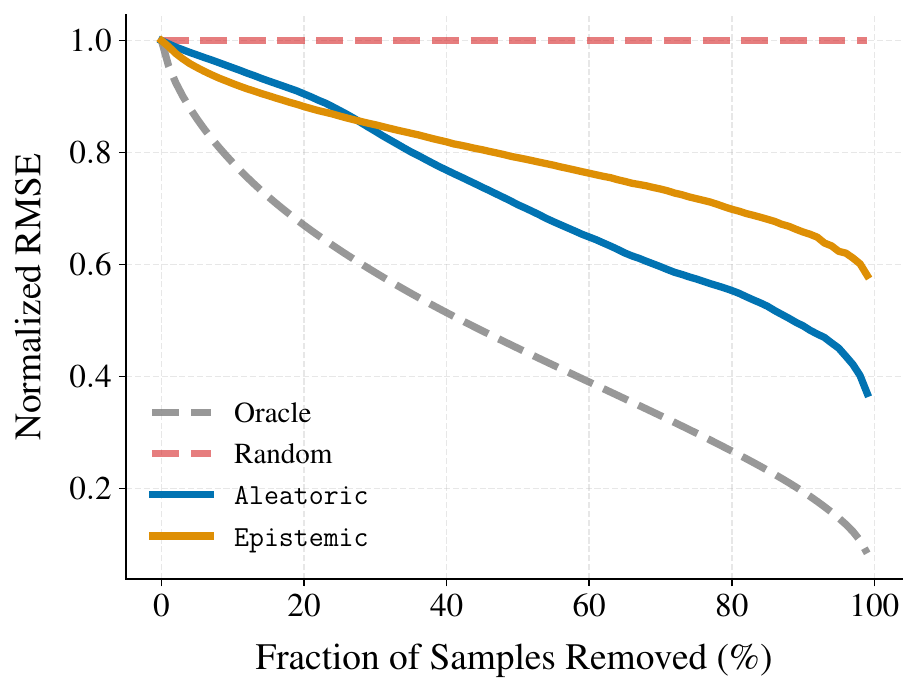}}
\\
\subfloat[Orientation (Simulation)]{\includegraphics[width=0.48\columnwidth]{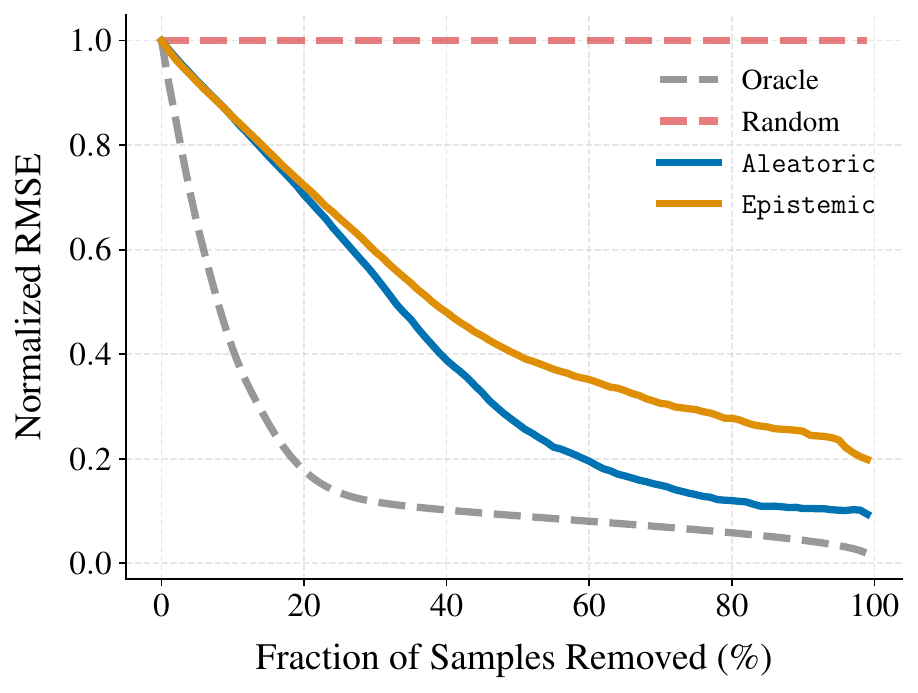}}
\hfil
\subfloat[Orientation (Real-world)]{\includegraphics[width=0.48\columnwidth]{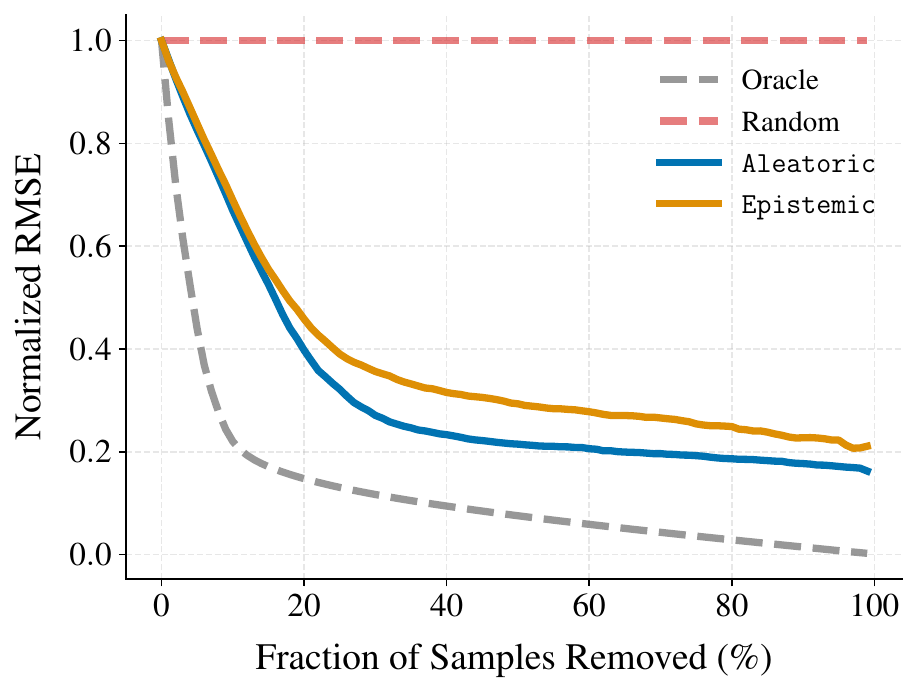}}
\caption{Sparsification curves for aleatoric and epistemic uncertainty. Samples are removed by decreasing uncertainty; oracle (grey dashed line) removal by true error gives the lower bound curve, while random removal (red dotted line) is the uninformative baseline. Both uncertainty sources outperform random removal across all panels.}
\label{fig:sparsification}
\end{figure}

We compare aleatoric and epistemic uncertainty on both the simulation test set (HM3D $D_{\text{Test}}^{\text{Sim}}$) and the real-world test set (CoViS-Net $D_{\text{Test}}^{\text{Real}}$). Table~\ref{tbl:uncertainty_source} reports the three metrics for position and orientation, and Fig.~\ref{fig:sparsification} shows the corresponding sparsification curves.

The consistent pattern across both regimes appears to reflect a structural difference between the two uncertainty sources. We hypothesize that, because the heteroscedastic GNLL head learns a sample conditioned variance tied to the regression loss, the same gradient that fits the mean may also push ambiguous samples (low FOV overlap, repetitive geometry, motion blur) toward larger predicted variance $\hat{\sigma}^2$. MC~Dropout, by contrast, samples function space perturbations from a data independent mask distribution, so its variance plausibly tracks parameter sensitivity at the input rather than the actual prediction error. Epistemic curves still sit clearly above the random baseline in Fig.~\ref{fig:sparsification}, so MC~Dropout does carry useful information about prediction reliability. However, parameter-sensitivity rankings are only weakly aligned with the visual cues that drive error, whereas aleatoric uncertainty preserves a stronger ordering, as shown in Table~\ref{tbl:uncertainty_source}, and is more robust to domain shift.

\subsection{Cross-Scene Generalization}\label{sec:cross_dataset}
To answer Q4, we evaluate cross-scene generalization by testing our HM3D trained model on the CoViS-Net real-world dataset~\cite{blumenkamp2025covis} without fine-tuning. Each sample is an $N$-tuple of synchronized monocular frames; we extend the original $N{=}3$ triplet protocol to $N \in \{2,3,4,5\}$ by enumerating all $N$-subsets of robots at each timestamp, yielding 56{,}032 samples. Both methods are evaluated on the same sample list and the same input frames, with each backbone's expected preprocessing, and pooled median errors are reported at each method's native granularity. High uncertainty predictions on the invisible subset are further rejected using Youden's index on $\hat{\sigma}_p$.

\begin{table}[!t]
\caption{%
  Cross-dataset generalization on CoViS-Net $D_{\text{Test}}^{\text{Real}}$ scenes without fine-tuning ($N \in \{2,3,4,5\}$). Position in meters, orientation in degrees. Visibility: $|\Delta\psi|<120^\circ$; filtering uses Youden's index on predicted uncertainty $\hat{\sigma}_p$. Best per column in \textbf{bold}.}
\label{tbl:crossdataset}
\centering
\setlength{\tabcolsep}{2pt}
\begin{tabular}{l|cc|cc|cc|cc}
\hline
 & \multicolumn{2}{c|}{All} & \multicolumn{2}{c|}{Visible} & \multicolumn{2}{c|}{Invisible} & \multicolumn{2}{c}{Invis.~Filt.} \\
Method & pos & ori & pos & ori & pos & ori & pos & ori \\
\hline
CoViS-Net~\cite{blumenkamp2025covis} & 0.632 & 7.934 & 0.396 & 6.219 & 0.730 & \textbf{8.624} & \textbf{0.297} & \textbf{5.705} \\
Ours                                  & \textbf{0.438} & \textbf{7.746} & \textbf{0.298} & \textbf{4.069} & \textbf{0.500} & 11.204 & 0.391 & 6.842 \\
\hline
\end{tabular}
\end{table}

Table~\ref{tbl:crossdataset} reports cross-scene generalization. For a fair comparison, CoViS-Net results are obtained by evaluating the authors' publicly released model weights on the same samples, using the same error definitions and the same visibility criterion, at the granularity native to each method as noted above. Our model outperforms CoViS-Net on position across all three unfiltered regimes (pooled position error: 0.438~m vs.\ 0.632~m, a $31\%$ reduction) and on orientation in the visible setting, while CoViS-Net retains a slight edge on invisible orientation. Uncertainty filtering on the invisible subset improves both models, and CoViS-Net ends ahead of us there on both position and orientation. Overall, both models trained purely on HM3D transfer to unseen real-world scenes, and the IVL formulation is competitive with the CoViS-Net pairwise baseline at $N \in \{2,3,4,5\}$.

\subsection{Onboard Inference Latency}\label{sec:latency}

Decentralized deployment requires every robot to run the estimator on its own compute, so we benchmark single robot latency on the NVIDIA Jetson AGX Orin carried in the physical experiments, over 100 timed runs after 20 warm-up runs with CUDA synchronization barriers. With the encoder exported to TensorRT and the GNN kept in TorchScript at FP16, a robot spends $5.7$\,ms in $f_{\mathrm{enc}}$ and $2.7$\,ms in $f_{\mathrm{gnn}}$, or $8.3$\,ms end to end, corresponding to roughly $120$\,Hz. CoViS-Net requires $9.4$\,ms under the same conditions. Only $f_{\mathrm{gnn}}$ depends on team size, and it stays within $2.5$--$3.0$\,ms from $N=2$ to $N=12$, still below the $3.8$\,ms CoViS-Net spends on a single pairwise message while returning pose for the whole formation rather than for one pair. Since the encoder runs once per image regardless of $N$, per-robot cost stays essentially flat over the measured range $N=2$ to $12$, which is what keeps the decentralized scheme practical as the team grows.

\subsection{Tunnel Traversal Experiment}
\label{sec:tunnel}

\begin{figure*}[!t]
\centering
\includegraphics[width=\textwidth]{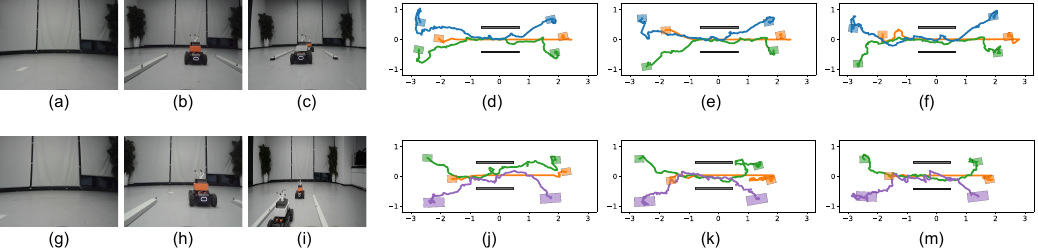}
\caption{\textbf{Homogeneous 3 RM and heterogeneous 2 RM+1 quadruped} setups. Left panels show in tunnel camera views. Right panels show top-down mocap trajectories traversing the tunnel. Colored lines are the mocap recorded paths of the mocap driven robot (RM$_0$, orange) and our network driven robots (RM$_1$, green; RM$_2$, blue; quadruped, purple). Gray rectangles mark the two parallel walls.\label{fig:tunnel_traj_combined}}
\end{figure*}

To answer Q5, we conduct two complementary sets of multi-robot formation morphing tunnel traversal experiments in a laboratory environment. In both setups, three robots approach a narrow tunnel in triangular formation, transition to a column formation at the entrance, traverse the corridor, and recover to triangular formation on exit. The first setup uses a homogeneous team of three RM wheeled robots (RM$_0$, RM$_1$, RM$_2$); the second replaces RM$_2$ with a Unitree Go2 quadruped, yielding a heterogeneous team with wheeled and quadruped robots that additionally stresses the model under platform diversity. The GNN model deployed here is trained entirely on real-world data. We collect this data by letting a single RM robot autonomously perform a random walk through our laboratory arena in an empty configuration, recording its onboard camera images together with the corresponding ground truth poses from the mocap system.

\subsubsection{Setting}
We use a hybrid control architecture in which RM$_0$ is trajectory driven in the tunnel, uses mocap to obtain its pose, and follows a preplanned centerline via a P controller, while the remaining two robots are GNN driven and rely solely on the GNN predictions for formation control. The tunnel length is $L=1.27$\,m and the corridor width is $W=0.72$\,m, which is approximately the combined width of three Robomasters. We report 12 trials for both homogeneous and heterogeneous setups.

\subsubsection{Results}
Fig.~\ref{fig:tunnel_traj_combined} shows mocap recorded trajectories for both setups. The team enters in triangular formation, contracts into the commanded column inside the tunnel, and reopens to triangular on exit; RM$_0$ with position feedback from mocap tracks its commanded centerline within $\sim$0.01\,m of lateral error throughout. All 24 trials successfully complete the traversal, with 21 of 24 collision-free: the homogeneous 3 RM team is collision-free in all 12 trials, while the heterogeneous wheeled+quadruped team is collision-free in 9 of 12 trials, with the remaining trials exhibiting only brief brushing contact.

Table~\ref{tbl:tunnel_errors} reports pose errors per phase for the two setups, from which three findings stand out. (i) Inside the tunnel, although the pose estimation error is around 0.2\,m (0.197 / 0.215\,m for the two setups), the GNN driven robots keep their lateral deviation from the corridor centerline within 0.065 / 0.103\,m. Since this lateral deviation is the quantity that determines whether a robot clears the walls, and its median stays below the available clearance, all 24 traversals complete. The 75th percentile ($0.159$\,m for the heterogeneous team) approaches the clearance left to the wider quadruped, which is consistent with the three brushing contacts reported above. (ii) The GNN driven quadruped attains 0.090\,m in tunnel lateral deviation, closely matching the wheeled robot despite its wider body and gait induced oscillations, which shows that the perception pipeline generalizes across platforms without fine-tuning. (iii) Predicted positional uncertainty $\hat{\sigma}_p$ rises from $\sim$0.064\,m in the open triangular phase to $\sim$0.086\,m in the tunnel, tracking scene difficulty.

\begin{table}[t]
\centering
\caption{Pose errors per phase for both tunnel setups, aggregated across 12 trials per setup. Each cell reports median / 75th percentile. ``Pose'' rows are Euclidean distance between GNN prediction and ground truth relative pose; ``Lateral'' rows are mocap measured lateral deviation from tunnel centerline.\label{tbl:tunnel_errors}}
\setlength{\tabcolsep}{3pt}
\begin{tabular}{l|l|c|c|c}
\hline
Setup & Phase & $e_p$ (m) & $e_\psi$ (deg) & $\hat{\sigma}_p$ (m) \\
\hline
\multirow{4}{*}{3 RM}
 & triangle (pose)    & 0.158 / 0.215 & 1.4 / 2.4 & 0.070 \\
 & transition (pose)  & 0.148 / 0.196 & 1.1 / 1.6 & 0.080 \\
 & in tunnel (pose)   & 0.197 / 0.248 & 1.3 / 2.0 & 0.081 \\
 & in tunnel (lateral)& 0.065 / 0.109 & 2.9 / 3.4 & 0.081 \\
\hline
\multirow{4}{*}{2 RM+Go2}
 & triangle (pose)    & 0.238 / 0.331 & 1.5 / 2.6 & 0.064 \\
 & transition (pose)  & 0.208 / 0.262 & 1.4 / 2.3 & 0.084 \\
 & in tunnel (pose)   & 0.215 / 0.267 & 1.6 / 2.4 & 0.086 \\
 & in tunnel (lateral)& 0.103 / 0.159 & 3.7 / 4.3 & 0.086 \\
\hline
\end{tabular}
\end{table}

Our experimental configuration is challenging in both setups: during training the model has never seen the tunnel walls used here or the robots within the formation; in column formation, each rear robot's environmental view is largely occluded by the robot in front of it; and the cameras face large gray-white motorized curtains that are nearly textureless and provide few distinctive visual cues. Notably, the two GNN driven robots receive no map, no waypoint, and no prior knowledge of the tunnel location: they clear the walls solely by holding formation from the estimated relative poses. RM$_0$ is the exception, following a preplanned centerline under mocap. Despite these difficulties, and despite consuming raw GNN output without temporal smoothing, both teams successfully traverse the 0.72\,m corridor (roughly the combined width of three RMs), underscoring the strong generalization of our perception pipeline and delivering feedback stable enough for direct closed loop control on both homogeneous and heterogeneous multi-robot teams.

\subsection{Generalization Summary}\label{sec:generalization_summary}
A recurring theme across the preceding experiments is that the learned estimator generalizes along four independent axes, none requiring fine-tuning. Axes (i), (ii) and (iv) are established by one model trained purely on HM3D; axis (iii) is established by the model trained on laboratory data, which is what the physical experiments deploy. (i)~\emph{Scene and sim-to-real}: the HM3D-trained model transfers directly to the CoViS-Net real-world benchmark at a pooled position error of $0.438$\,m, $31\%$ below the CoViS-Net baseline (Section~\ref{sec:cross_dataset}). (ii)~\emph{Team size}: although trained only on $N\in\{2,\dots,5\}$, the same model holds at $0.29$\,m position error on the entirely unseen $N=6$ and $N=7$ (Section~\ref{sec:pose_estimation_accuracy}). (iii)~\emph{Platform}: the laboratory-trained model sees only wheeled-robot imagery during training, yet the quadruped it drives matches the wheeled robots' in-tunnel lateral deviation to within $0.090$\,m despite its wider body and gait-induced motion (Section~\ref{sec:tunnel}). (iv)~\emph{Team composition and connectivity}: estimation depends only weakly on which teammate is lost (worst-node index $1.24$) and degrades gracefully under communication sparsification (Section~\ref{sec:robustness}). That the same architecture and training objective, applied to two disjoint data sources, spans all four axes and closes the loop for real formation control across $24$ physical trials, is the central evidence that the learned IVL representation is transferable rather than tuned to a single operating point.

\section{Conclusion}
We presented a decentralized, vision-only relative pose estimation framework for multi-robot formation control based on the IVL, a non-physical reference frame implicitly learned by a Transformer based GNN. Using only monocular images and inter-robot communication, the model estimates each robot's 6-DoF pose relative to the IVL while providing both aleatoric and epistemic uncertainty. Experiments show strong cardinality and scene generalization across 2--7 robots, effective aleatoric uncertainty under simulation and real-world tests, direct transfer from HM3D to real-world benchmarks without fine-tuning, and graceful degradation with no single point of failure under robot dropout and communication sparsification. In physical tunnel traversal experiments, the same framework provides stable closed-loop feedback for both homogeneous and heterogeneous robot teams, enabling traversal of a narrow 0.72\,m corridor, thereby demonstrating both high precision and practical generalizability.

Several limitations remain. The current framework performs single frame inference and does not exploit temporal consistency, which could further improve pose stability. A sim-to-real gap also remains under real-world domain shift. Orientation estimation is also more sensitive than position to communication sparsification, since it relies more heavily on cross-robot cues; training with random graph augmentation is a natural remedy. Finally, since this work focuses on perception and relative localization, future efforts will build on this framework toward integrated obstacle avoidance, planning, and navigation for multi-robot formations.

%

\bibliographystyle{IEEEtran}
\bibliography{refs}

\end{document}